\documentclass[letterpaper]{article} 
\usepackage{aaai2026}
\nocopyright
\usepackage{times}  
\usepackage{helvet}  
\usepackage{courier}  
\usepackage[hyphens]{url}  
\usepackage{graphicx} 
\usepackage{natbib}  
\usepackage{caption} 
\usepackage{algorithm}
\usepackage{algorithmic}

\usepackage{newfloat}
\usepackage{listings}
\DeclareCaptionStyle{ruled}{labelfont=normalfont,labelsep=colon,strut=off} 
\floatstyle{ruled}
\newfloat{listing}{tb}{lst}{}
\floatname{listing}{Listing}
\usepackage{booktabs}

\title{Not Just After One: Sleep-Inspired Replay Prevents Catastrophic Forgetting After Sequential Tasks}

\author{
Anthony Bazhenov\textsuperscript{\rm 1}
Jean Erik Delanois\textsuperscript{\rm 2}
Giri P. Krishnan\textsuperscript{\rm 3}
}
\affiliations{
    \textsuperscript{\rm 1}
    Khoury College of Computer Sciences, Northeastern University, Boston, MA \\ 
    \textsuperscript{\rm 2}
    Department of Medicine, University of California, San Diego, La Jolla, CA, USA \\
    \textsuperscript{\rm 3}
    ARTISAN, Georgia Institute of Technology, Atlanta, GA USA
    \\
    anthbazhenov@gmail.com, jdelanois@ucsd.edu, giri@gatech.edu}
\usepackage{bibentry}

\begin{document}

\maketitle

\begin{abstract}
One of the critical limitations of artificial neural networks is their lack of ability to continually learn: training on new tasks often leads to  interference and forgetting of the previous ones. While several algorithms have been proposed to protect old memories from interference, they are typically applied during or immediately after each new episode of training. In contrast, humans and animals can learn continuously, acquiring multiple new memories during active learning before consolidating all of them into long-term storage. Here we show that multiple new tasks can be trained sequentially before an unsupervised sleep-like replay phase is applied to partially restore performance across all previously learned tasks. Our study further suggests that task-specific information remains resilient to new training but decays gradually as network is trained on new tasks. These findings point to novel principles for developing a broad range of continual learning AI solutions.
\end{abstract}


\section{Introduction}

Artificial neural networks (ANNs), including LLMs, suffer from "catastrophic forgetting," whereby they achieve optimal performance on newer tasks at the expense of performance on previously learned tasks \cite{french1999catastrophic, hayes2021replay,luo2024}. A powerful approach to prevent catastrophic forgetting is rehearsal \cite{kemker2018measuring} combining
previously learned data with new task data \cite{robins1995catastrophic}. While its disadvantage is the need to either preserve old data or regenerate them using complex models, an unsupervised form of rehearsal - Sleep Replay Consolidation (SRC) \cite{Tadros_NC2022} - was proposed to mimic biological sleep.
Importantly, all these methods are usually applied either during or immediately after each new episode of training.

Neuroscience research suggests that memory replay during sleep plays an important role in forming stable long-term memory representations \cite{stickgold2005sleep}.
Co-replay of recently learned memories along with relevant old ones during sleep 
helps form stable, orthogonal memory representations and effectively divides network resources between multiple memories \cite{gonzalez2020can}. However, the biological brain can acquire multiple, possibly overlapping, memories before a single episode of night sleep consolidates all of them. How is this achieved?

In this study, we demonstrate that SRC can rescue older tasks impaired by multiple episodes of new training. Although performance on older tasks progressively degraded as the number of tasks trained before SRC was applied increased, enough information was preserved to allow unsupervised sleep-like replay to restore significant performance on these memories.

\section{Algorithm}
To examine how post-training sleep influences networks in sequential-learning settings, we used the MNIST dataset of handwritten digits (0–9), the Fashion-MNIST dataset of Zalando articles and the CIFAR-10 dataset — all standard benchmarks for sequential and continual-learning experiments \cite{lecun1998gradient, xiao2017fashion}.

In this protocol, tasks were presented in order, with sleep introduced only after the final task in each simulation. For example, in one simulation the network was trained first on Task 1, followed by Task 2, and then entered a sleep stage. In another simulation the network was trained on Task 1, Task 2, and Task 3 before sleep was applied. This protocol was repeated for up to all 5 tasks, allowing us to evaluate whether a terminal sleep stage could enhance retention and integration of knowledge across all tasks in the sequence.


All simulations employed the same network architecture and training parameters. For MNIST and FMNIST used a fully connected feedforward network with two hidden layers of 1200 units each, followed by a 10-unit classification output layer that was trained  via backpropagation and stochastic gradient descent. Hidden layers used ReLU nonlinearities. To promote generalization and robustness, training incorporated dropout in the hidden layers and a binary cross entropy loss, with weights updated using a standard stochastic gradient descent optimizer. No biases were included for any neurons, simplifying the eventual conversion to spiking dynamics. 
\begin{figure}[h]
    \centering
    \includegraphics[width=0.43\textwidth]{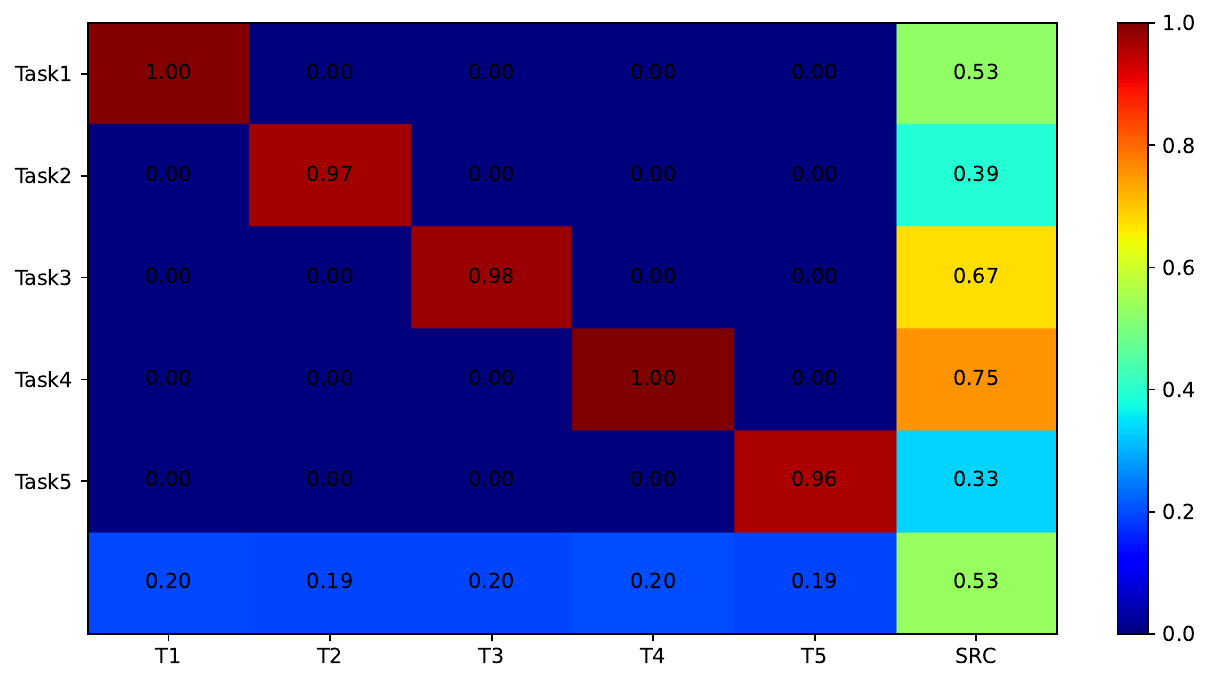}
    \includegraphics[width=0.43\textwidth]
    {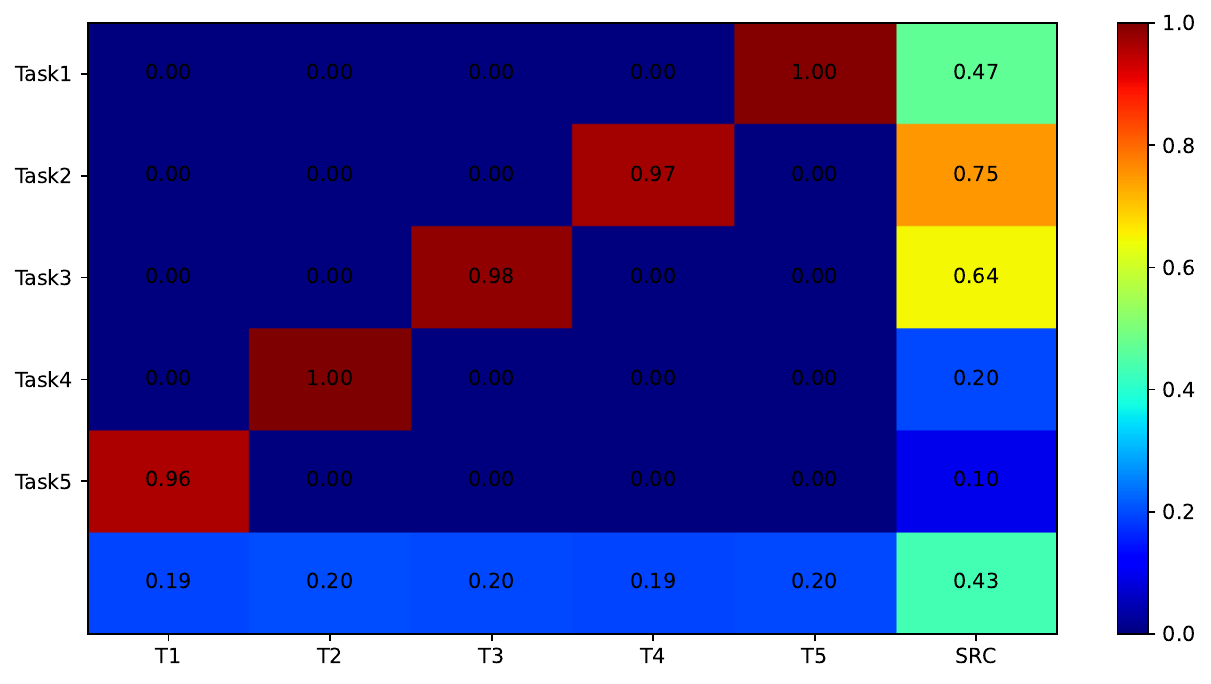}
    \caption{Applying sleep after five tasks trained in a sequence partially recovers performance for MNIST data. Sleep phase (SRC) was implemented at the end after all tasks training. Each column shows performance for all tasks after either new task training or SRC as labeled below. 
    }
    \label{fig:matrix}
\end{figure}
For CIFAR-10, we used a compact CNN with two 3×3 convolution layers (3→50 and 50→50, no bias), each followed by ReLU and 2×2 max-pooling, then a fully connected head consisting of a flatten operation and two linear layers (1800→1200 with ReLU, 0.3 dropout, then 1200→10). The two linear layers were initialized with Xavier uniform initialization, and training optionally supported loading pretrained weights and selectively freezing specific layers via per-layer requires\_grad flags. During each forward pass, intermediate activations (input, after each conv/pool block, after the hidden FC, and the final logits) were stored and exposed for logging.

Sleep Replay Consolidation (SRC) algorithm was implemented as described in \cite{Tadros_NC2022}. 
After training ANN with backpropagation, the unsupervised sleep stage SRC is applied.
To convert the ANN to a SNN, the network's activation function is replaced by a Heaviside function and weights are scaled by the layer-wise activation maximum observed on the training dataset, as suggested by \cite{diehl2015fast}, to ensure the network maintains reasonable firing activity. The SRC phase starts with a forward pass where Poisson distributed spike trains reflective of input pixel averages are generated and fed to the input layer in order to propogate activity (spiking behavior) across the network. Following the forward pass, a backward pass is executed to update synaptic weights. To modify network connectivity during sleep we use an unsupervised simplified Hebbian type learning rule which is implemented as follows: a weight is increased between two nodes when both pre- and post-synaptic nodes are activated (i.e., input exceeds the Heaviside activation function threshold), a weight is decreased between two nodes when the post-synaptic node is activated but the pre-synaptic node is not (in this case, another pre-synaptic node is responsible for activity in the post-synaptic node). After running multiple steps of this unsupervised learning during sleep, the final weights are rescaled again (simply by removing the original scaling factor), the Heaviside-type activation function is replaced by ReLU, and testing or further supervised training on new data is performed.

\begin{figure}[H]
    \centering
    \includegraphics[width=0.45\textwidth]{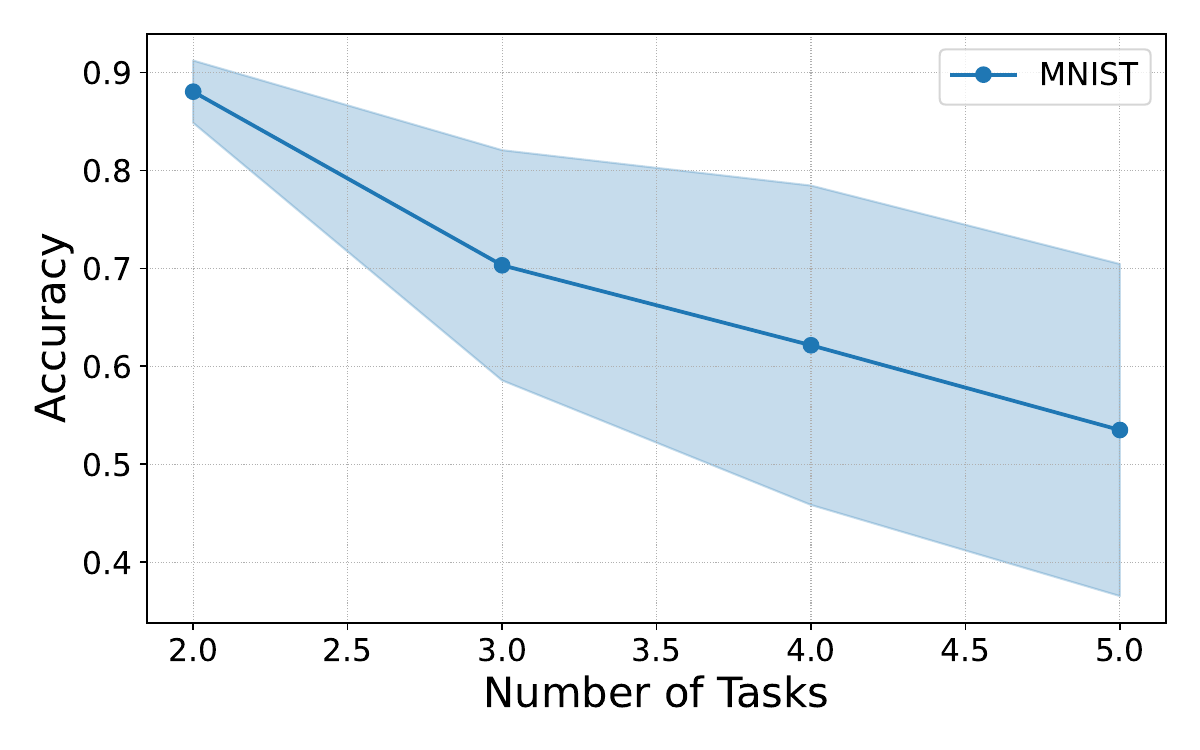}
    \includegraphics[width=0.45\textwidth]{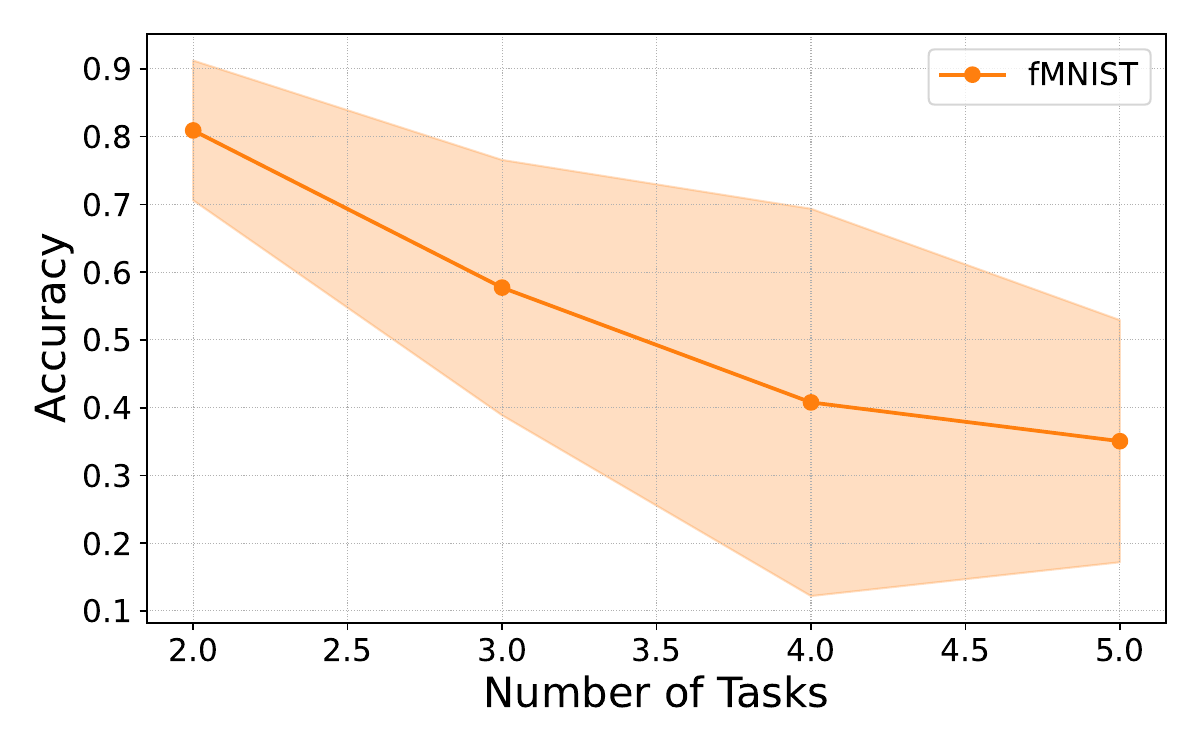}
    \includegraphics[width=0.45\textwidth]{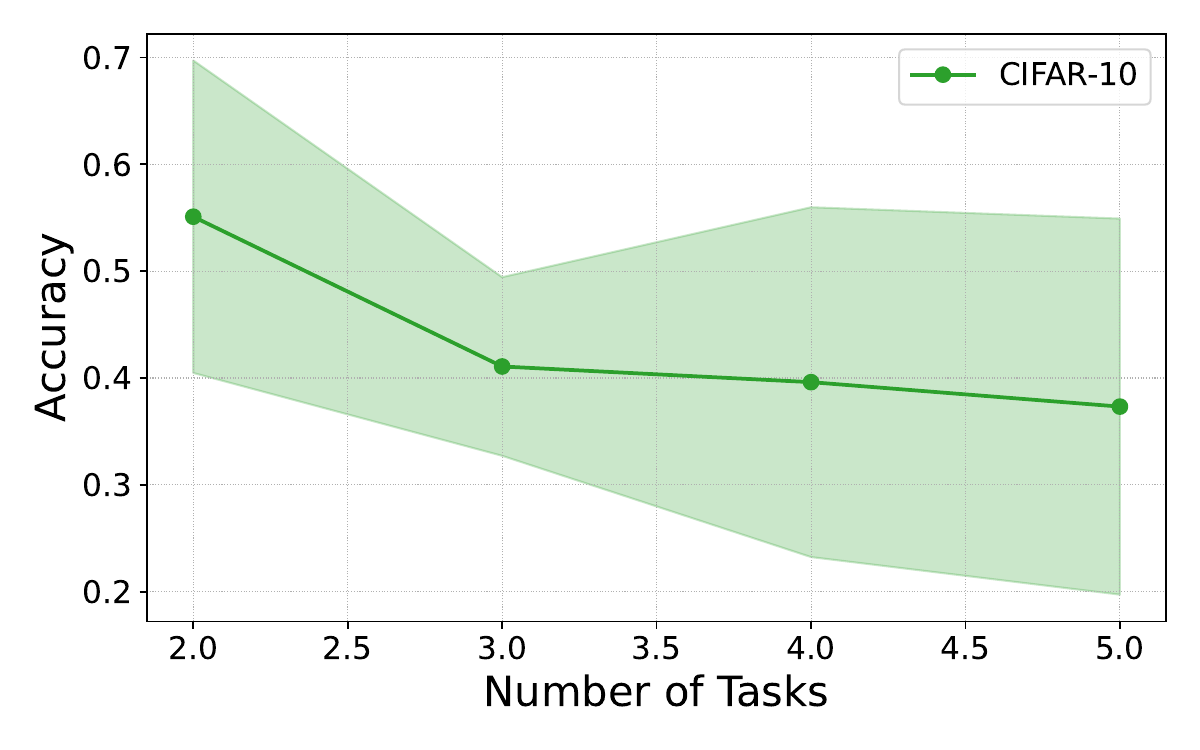}
    \caption{Mean ± SD of accuracy vs. number of tasks trained. SRC was applied only once, after sequences of 2, 3, 4, or 5 tasks. Accuracy was averaged across all tasks in each sequence and across 10 trials; error bars show ±1 standard deviation.}
  \label{fig:mean}
\end{figure}

\section{Results}

Previous studies reported that SRC applied after each new task can prevent catastrophic forgetting in feedforward \cite{Tadros_NC2022} and recurrent \cite{Kubo2025} networks. Here we tested whether applying SRC only once, after a full sequence of tasks, can still produce performance recovery. We found that training each new task caused near-complete forgetting of earlier tasks. However, a single SRC episode applied after a sequence of tasks recovered a significant portion of accuracy across previously trained tasks (Figure \ref{fig:matrix}). The amount of recovery varied across tasks: later tasks in the sequence did not necessarily show the best recovery. Changing task order affected individual task accuracy after SRC but not the  mean performance across all tasks (Figure \ref{fig:matrix}).

We next systematically analyzed how the number of tasks trained before SRC affected end performance (Figure \ref{fig:mean}). Increasing the number of tasks trained before SRC led to lower average performance; however, the decrease was gradual, suggesting that training additional tasks progressively increased interference with earlier tasks. Interestingly, for CIFAR-10, performance dropped substantially when the number of tasks increased from 2 to 3, but then largely saturated with further increases in the number of tasks (Figure \ref{fig:mean}, bottom).

\begin{figure}[h]
    \centering
    \includegraphics[width=0.45\textwidth]{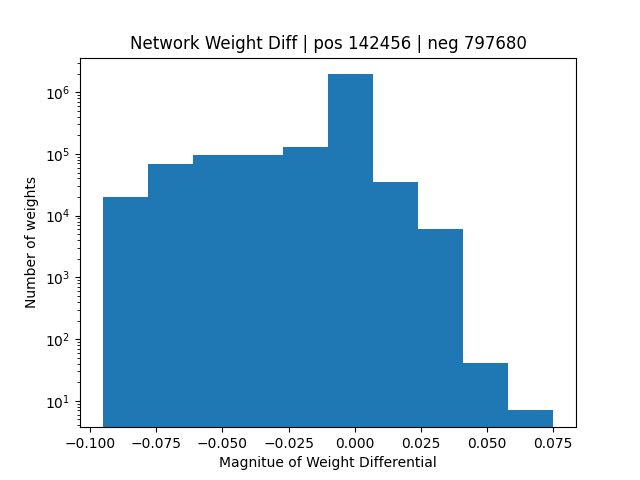}
    \includegraphics[width=0.45\textwidth]{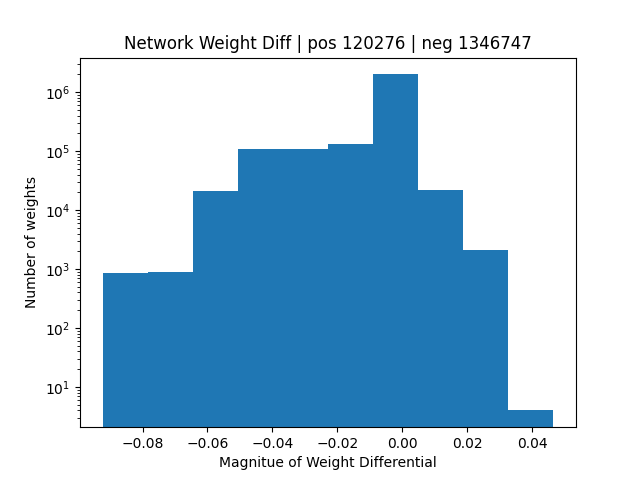}
    \caption{Weight difference distributions for MNIST (top) and FMNIST (bottom). Plots highlight that many synapses decrease in strength implying performance benefits are a result of suppressing incorrect signals.  }
    \label{fig:srcWeightDiffDist}
\end{figure}

In an attempt to gain further insight as to why SRC was capable of improving  under trained model performance, we analyzed the sleep stage itself. We first measured network activity by simply looking at instantaneous layerwise firing rates. It can be observed that activity decreased from first hidden to output layer, suggesting that main benefits of SRC were  due to hidden
layer modifications. Additionally the magnitude of weight modifications was examined. Figure \ref{fig:srcWeightDiffDist} shows a histogram of the cumulative weight perturbation each synapse received during sleep. Although a small number of critical synapses showed an increase strength (Figure \ref{fig:srcWeightDiffDist} - positive weight deltas) most synapses decreased their sensitivity (Figure \ref{fig:srcWeightDiffDist} - vast number of synapses with weight deltas less than 0). Taken together, these results imply that SRC improves feature representations by making hidden-layer neurons less sensitive to certain (overlapping) stimulus features. They further suggest that SRC selectively suppresses many synapses while maintaining a few, thereby reducing overlap between tasks and leading the model to focus on the most predictive features for each task.

\section{Conclusions}

Our study suggests forgetting in ANNs is a gradual process. Although the last task in a sequence often grabs most of the performance, which can look like complete loss of earlier information, that information is still present and can be recovered by sleep-like unsupervised replay. Importantly, damage to earlier tasks was not cumulative: earlier tasks could recover a similar or even higher amount of accuracy as later ones. This implies damage depends more on interference between tasks (as suggested in \cite{Saxena2022PNAS}); tasks with less overlap show better recovery regardless of their position in the training sequence.

\section{Funding}
Supported by NSF (grants 2209874 and 2223839) and NIH (grant RFNS132913).

\bibliography{references,references-NC} 

\end{document}